\pdfoutput=1
\documentclass[letterpaper, 10 pt, conference]{ieeeconf}
\IEEEoverridecommandlockouts
\usepackage[T1]{fontenc}
\usepackage[utf8]{inputenc}
\usepackage{times}   
\usepackage{amsmath,amssymb,bm}
\usepackage{graphicx}
\usepackage{booktabs}
\usepackage{multirow}
\usepackage{array}
\usepackage{tabularx}
\usepackage{threeparttable}
\usepackage{makecell}
\usepackage{xcolor}
\usepackage{tikz}
\usetikzlibrary{positioning,arrows.meta,calc}
\usepackage{cite}
\usepackage{url}
\usepackage{balance}

\graphicspath{{figures/}}
\newif\ifprovisionalcohort
\newif\ifprelimprovisional

\newcommand{\nPw}{ten}

\newcommand{\nCohortw}{ten}

\newcommand{\nFullw}{six}

\newcommand{\nChanExclw}{seven}
\newcommand{\nChan}{57}
\newcommand{\nChanFeat}{114}
\newcommand{\nChanFeatThree}{171}
\newcommand{\nFrameFeat}{195}

\newcommand{\chanExclList}{EMG\_25, EMG\_34, EMG\_54, EMG\_61--EMG\_64}
\newcommand{\nTargets}{60}
\newcommand{\nDonors}{nine}
\newcommand{\nFolds}{ten}
\newcommand{\nSignFlips}{$2^{10}=1024$}
\newcommand{\nRecEval}{540}

\newcommand{\nCohortRecords}{180}
\newcommand{\nCohortTargets}{60}
\newcommand{\nCohortDonors}{nine}

\newcommand{\nMinP}{0.002}
\newcommand{\nMinPfrac}{2/1024}

\newcommand{\recordsPerParticipantHeld}{54}
\newcommand{\vRcRtwo}{0.7222}

\newcommand{\vRcRmse}{6.681}

\newcommand{\vRcMae}{5.146}
\newcommand{\vRcR}{0.896}

\newcommand{\vRcParams}{76,534}
\newcommand{\vGruParams}{39,478}

\newcommand{\vImuOnlyRtwo}{0.212}
\newcommand{\vGruEmgRtwo}{0.710}
\newcommand{\vFdRtwo}{0.509}
\newcommand{\vGruRtwoThree}{0.729}
\newcommand{\vTcnDR}{\ensuremath{-0.0065}}
\newcommand{\vTcnDRThree}{\ensuremath{-0.006}}

\newcommand{\vTcnDRCi}{\ensuremath{[-0.0169,\,+0.0022]}}

\newcommand{\vImuDR}{\ensuremath{+0.0022}}
\newcommand{\vImuDRThree}{\ensuremath{+0.002}}

\newcommand{\vImuDRCi}{\ensuremath{[-0.0450,\,+0.0476]}}

\newcommand{\vTdDR}{\ensuremath{+0.2107}}
\newcommand{\vTdDRThree}{\ensuremath{+0.211}}

\newcommand{\vTdDRCi}{\ensuremath{[+0.1645,\,+0.2638]}}

\newcommand{\vTdDRmseTwo}{\ensuremath{-2.84}}

\newcommand{\vTdImpRw}{ten of ten}

\newcommand{\vSToRtwo}{0.7222}
\newcommand{\vSToRmse}{6.681}

\newcommand{\vSToRtwoThree}{0.722}

\newcommand{\vSRpRtwo}{0.7400}

\newcommand{\vSPoRtwo}{0.7880}
\newcommand{\vSPoRmse}{5.631}

\newcommand{\vSPoRtwoThree}{0.788}

\newcommand{\vTotDR}{\ensuremath{+0.0658}}
\newcommand{\vTotDRThree}{\ensuremath{+0.066}}
\newcommand{\vTotDRCi}{\ensuremath{[+0.0361,\,+0.0955]}}

\newcommand{\vTotDRmse}{\ensuremath{-1.049}}

\newcommand{\vTotDRmseCi}{\ensuremath{[-1.462,\,-0.649]}}

\newcommand{\vTotImpRw}{ten of ten}

\newcommand{\vInfDR}{\ensuremath{+0.0480}}
\newcommand{\vInfDRThree}{\ensuremath{+0.048}}
\newcommand{\vInfDRCi}{\ensuremath{[+0.0166,\,+0.0790]}}

\newcommand{\vInfDRmse}{\ensuremath{-0.736}}

\newcommand{\vInfDRmseCi}{\ensuremath{[-1.122,\,-0.339]}}

\newcommand{\vInfImpRw}{seven of ten}
\newcommand{\vInfImpRmsew}{nine of ten}

\newcommand{\vBudDR}{\ensuremath{+0.0178}}
\newcommand{\vBudDRThree}{\ensuremath{+0.018}}
\newcommand{\vBudDRmse}{\ensuremath{-0.313}}

\newcommand{\vBudShareR}{27\%}
\newcommand{\vBudShareRmse}{30\%}
\newcommand{\vSdDonorDR}{\ensuremath{-0.0601}}

\newcommand{\vSdDonorDRCi}{\ensuremath{[-0.0970,\,-0.0291]}}
\newcommand{\vSdDonorDRmse}{\ensuremath{+0.630}}
\newcommand{\vSdDonorDRmseCi}{\ensuremath{[+0.244,\,+1.078]}}

\newcommand{\vSdDonorPholm}{0.023}

\newcommand{\vSdDonorWorstR}{0.189}

\newcommand{\vSsDonorDR}{\ensuremath{+0.0365}}

\newcommand{\vSsDonorDRCi}{\ensuremath{[+0.0153,\,+0.0605]}}

\newcommand{\vSsDonorPholm}{0.039}

\newcommand{\vOtherDonorMaxAbs}{0.03}
\newcommand{\vRecvFwDR}{\ensuremath{-0.026}}

\newcommand{\vRecvSwDR}{\ensuremath{-0.012}}

\newcommand{\vRecvSdDR}{\ensuremath{+0.005}}

\newcommand{\vRecvStwDR}{\ensuremath{+0.024}}

\newcommand{\vRecvMeanDR}{\ensuremath{+0.001}}
\newcommand{\vDirectedPairs}{30}
\newcommand{\vDirectedTests}{60}

\newcommand{\vDurDR}{\ensuremath{-0.0182}}
\newcommand{\vDurDRCi}{\ensuremath{[-0.0352,\,-0.0032]}}
\newcommand{\vDurDRP}{0.047}

\newcommand{\vDurDRmse}{\ensuremath{+0.220}}
\newcommand{\vDurDRmseCi}{\ensuremath{[+0.073,\,+0.365]}}

\newcommand{\vDurPholm}{0.051}
\newcommand{\vDurVif}{4.4}
\newcommand{\vPtwoPoolDR}{\ensuremath{+0.0480}}

\newcommand{\vPtwoPoolNegPct}{40\%}

\newcommand{\vPtwoFixedDR}{\ensuremath{+0.0408}}

\newcommand{\vPtwoFixedNegPct}{38\%}

\newcommand{\vPtwoGateDR}{\ensuremath{+0.0024}}
\newcommand{\vPtwoGateDRCi}{\ensuremath{[-0.0228,\,+0.0195]}}

\newcommand{\vPtwoGateSelN}{20}

\newcommand{\vPtwoOracleDR}{\ensuremath{+0.0809}}
\newcommand{\vPtwoOracleDRCi}{\ensuremath{[+0.0534,\,+0.1124]}}
\newcommand{\vPtwoOracleDRThree}{\ensuremath{+0.081}}

\newcommand{\vPtwoOracleSelN}{53}

\newcommand{\vPtwoOracleRatio}{1.7}

\newcommand{\vSWcRtwo}{0.8100}
\newcommand{\vSWcRmse}{5.220}

\newcommand{\vSWcRtwoThree}{0.810}

\newcommand{\vSPcRtwoThree}{0.210}

\newcommand{\vBestStateName}{W with cross-subject pretraining}
\newcommand{\vGtDR}{\ensuremath{+0.0525}}
\newcommand{\vGtDRThree}{\ensuremath{+0.053}}
\newcommand{\vGtDRCi}{\ensuremath{[+0.0189,\,+0.0872]}}

\newcommand{\vGtDRmse}{\ensuremath{-0.871}}
\newcommand{\vGtDRmseCi}{\ensuremath{[-1.292,\,-0.455]}}

\newcommand{\vGtPholmR}{0.070}
\newcommand{\vGtPholmRmse}{0.018}

\newcommand{\vGtImpRw}{eight of ten}

\newcommand{\vGwDR}{\ensuremath{+0.0277}}
\newcommand{\vGwDRThree}{\ensuremath{+0.028}}
\newcommand{\vGwDRCi}{\ensuremath{[+0.0072,\,+0.0486]}}

\newcommand{\vGwDRmse}{\ensuremath{-0.385}}
\newcommand{\vGwDRmseCi}{\ensuremath{[-0.643,\,-0.152]}}

\newcommand{\vGwPholmRmse}{0.018}

\newcommand{\vIsDR}{\ensuremath{-0.0249}}

\newcommand{\vIsDRCi}{\ensuremath{[-0.0542,\,+0.0005]}}

\newcommand{\vIsDRmse}{\ensuremath{+0.486}}
\newcommand{\vIsDRmseCi}{\ensuremath{[+0.224,\,+0.758]}}

\newcommand{\vIsPholmRmse}{0.018}

\newcommand{\vIsNotImpRmsew}{nine of ten}

\newcommand{\vGtStwDR}{\ensuremath{+0.133}}

\newcommand{\vGtFwDR}{\ensuremath{-0.006}}

\newcommand{\vGtSsDR}{\ensuremath{+0.077}}

\newcommand{\vGtSwDR}{\ensuremath{-0.010}}

\newcommand{\vGtKneeDR}{\ensuremath{+0.105}}

\newcommand{\vGtKneePholm}{0.018}

\newcommand{\vGtHipDR}{\ensuremath{+0.002}}

\newcommand{\vRecGtR}{0.440}
\newcommand{\vRecGtRTwo}{0.44}
\newcommand{\vRecGtRCi}{\ensuremath{[0.169,\,0.721]}}

\newcommand{\vRecGtRPholm}{0.067}

\newcommand{\vRecGwRTwo}{0.27}
\newcommand{\vRecGtRmse}{0.592}

\newcommand{\vRecGtRmseCi}{\ensuremath{[0.274,\,0.804]}}

\newcommand{\vRecGtRmsePholm}{0.021}

\newcommand{\vRecGwRmseTwo}{0.50}
\newcommand{\vLopoGtPholmMin}{0.23}

\newcommand{\vLopoGwWorseP}{0.039}

\newcommand{\vPthreeTAlwaysDR}{\ensuremath{+0.0525}}

\newcommand{\vPthreeTAlwaysNeg}{20}
\newcommand{\vPthreeTAlwaysNegPct}{33\%}

\newcommand{\vPthreeTGateDR}{\ensuremath{+0.0362}}

\newcommand{\vPthreeTGateDRCi}{\ensuremath{[+0.0115,\,+0.0618]}}
\newcommand{\vPthreeTGateDRmse}{\ensuremath{-0.590}}

\newcommand{\vPthreeTGateSelN}{28}
\newcommand{\vPthreeTGatePos}{20}
\newcommand{\vPthreeTGateNeg}{8}

\newcommand{\vPthreeTOracleDR}{\ensuremath{+0.0675}}
\newcommand{\vPthreeTOracleDRThree}{\ensuremath{+0.068}}
\newcommand{\vPthreeTOracleDRCi}{\ensuremath{[+0.0400,\,+0.0959]}}

\newcommand{\vPthreeWAlwaysDR}{\ensuremath{+0.0277}}

\newcommand{\vPthreeWGateDR}{\ensuremath{+0.0086}}

\newcommand{\vPthreeWGateDRCi}{\ensuremath{[-0.0048,\,+0.0250]}}

\newcommand{\vPthreeWOracleDR}{\ensuremath{+0.0419}}

\newcommand{\vPthreeTOracleRatio}{1.3}
\newcommand{\vInfNonImpList}{P04, P08 and P09}
\newcommand{\vInfNonImpN}{three}

\newcommand{\vImuMin}{\ensuremath{-0.120}}
\newcommand{\vImuMax}{\ensuremath{+0.124}}
\newcommand{\vTcnMin}{\ensuremath{-0.044}}
\newcommand{\vTcnMax}{\ensuremath{+0.016}}

\newcommand{\vMaxVif}{33.8}

\newcommand{\vSeedsPairStage}{one seed per network (20260718)}

\newcommand{\vPtwoPoolNegPctText}{40\%}
\newcommand{\vPtwoFixedNegPctText}{38\%}

\provisionalcohortfalse
\prelimprovisionalfalse

\newcommand{\dg}{\ensuremath{^{\circ}}}
\newcommand{\R}{\ensuremath{R^{2}}}
\newcommand{\dR}{\ensuremath{\Delta R^{2}}}
\newcommand{\dRMSE}{\ensuremath{\Delta\mathrm{RMSE}}}
\newcommand{\GT}{\ensuremath{G_{\mathrm{T}}}}
\newcommand{\GW}{\ensuremath{G_{\mathrm{W}}}}
\newcommand{\Istate}{\ensuremath{I_{\mathrm{state}}}}
\newcommand{\rctcn}{RC-TCN--GRU}
\newcommand{\ci}[2]{[#1,\,#2]}
\newcolumntype{L}[1]{>{\raggedright\arraybackslash}p{#1}}
\newcolumntype{C}[1]{>{\centering\arraybackslash}p{#1}}

\colorlet{cOwn}{blue!60!black}
\colorlet{cOther}{teal!70!black}
\colorlet{cCross}{orange!80!black}
\colorlet{cGray}{black!70}
\tikzset{
  blk/.style={draw=#1, fill=#1!12, rounded corners=2pt, thick, align=center, inner sep=3pt, font=\scriptsize, text width=2.9cm},
  blk/.default=black,
  arr/.style={-{Latex[length=1.6mm]}, thick, draw=black!70},
  note/.style={font=\fontsize{6}{7}\selectfont, align=left, text=black!70},
}

\title{\LARGE \bf Continuous Lower-Limb Joint-Angle Estimation from sEMG and IMU: The Conditional Value of Cross-Activity and Cross-Subject Data under Update-Matched Controls}

\author{Jiaying Xu$^{1}$, Guangyuan Chen$^{2}$, Buyun Gao$^{1}$, and Zhen Dong$^{3}$%
\thanks{This work has been submitted to the IEEE for possible publication. Copyright may be transferred without notice, after which this version may no longer be accessible.}%
\thanks{$^{1}$J. Xu and B. Gao are with Conamur, Palo Alto, CA, USA. Corresponding author: Jiaying Xu, {\tt\small alice@conamur.ai}.}%
\thanks{$^{2}$G. Chen is with the Institute of Neuroscience, Chinese Academy of Sciences, Shanghai, China, {\tt\small chengy2023@ion.ac.cn}.}%
\thanks{$^{3}$Z. Dong is with the University of California, Santa Barbara, CA, USA, {\tt\small zhen@ucsb.edu}.}}

\makeatletter
\@ifundefined{keywords}{%
  \newenvironment{keywords}{\par\vspace{2pt}\noindent\small\textbf{Index Terms}---}{\par\vspace{4pt}}%
}{}
\makeatother

\begin{document}

\maketitle
\thispagestyle{empty}
\pagestyle{empty}

\begin{abstract}
Intent-driven exoskeleton control needs a continuous, causal estimate of the wearer's joint kinematics and an answer to a second question: whose data should train the estimator for a new user? Borrowing the user's recordings of other activities, or other users' recordings of the same activity, is usually evaluated as ``more data'' versus ``less data,'' which confounds the information a source carries with the optimizer updates it adds and hides negative transfer behind averages. We separate these effects for hip, knee, and ankle angle estimation from high-density sEMG and an IMU across six locomotion activities on all \nCohortw{} participants of a public dataset. A compact residual causal TCN--GRU decoder is frozen, and an \emph{update-matched replay control} prices every added source against the same number of updates on the user's own data. A user's own other activities raise \R{} from \vSToRtwoThree{} to \vSPoRtwoThree{}, of which $\vInfDRThree$ is information beyond replay; stair descent is a reliably harmful donor and sit--stand--walk a reliably beneficial one. Other users' same-activity data help while the user has only the target activity ($\GT{}=\vGtDRThree$) and half as much once own other-activity data exist ($\GW{}=\vGwDRThree$); at full calibration, donor value collapses after one to two target records. Per-target gains from the two sources correlate, and leave-one-participant-out gates recover most of the cross-subject value only at low calibration. Data value is hierarchical, state-dependent, and directional; the bottleneck is deciding, without test labels, when to transfer.

\end{abstract}


\section{INTRODUCTION}
\label{sec:intro}
Lower-limb exoskeletons reduce the metabolic cost of walking \cite{collins2015reducing}, and assistance is now personalized by human-in-the-loop optimization of controller parameters \cite{slade2022personalizing}, by estimating the user's joint moments with a neural network \cite{molinaro2024estimating}, or by driving the device from the wearer's own neuromuscular intent \cite{gordon2007learning}. All of these, and the co-adaptive controllers that would let decoder and wearer adapt jointly \cite{madduri2023biosignal}, need a continuous, low-latency estimate of intended joint motion. Surface electromyography (sEMG) carries anticipatory information, inertial sensing (IMU) provides kinematic context, and neural-network regressors---shallow networks for simultaneous lower-limb joints \cite{zhang2012semg}, recurrent and convolutional--recurrent networks \cite{xia2018emg,bao2021cnnlstm}, transfer-learning-based recurrent models for the knee \cite{gautam2020myonet}---are the standard tools for mapping them to continuous joint angles. Reported errors depend on protocol and units (hip, knee, and ankle RMSE below 6\dg{} in treadmill walking \cite{zhang2012semg}; a normalized knee MAE of 8.1\% \cite{gautam2020myonet}); inter-subject variability remains the recognized obstacle, electrode placement, skin impedance, and recruitment strategy all shifting the sensor-to-output mapping \cite{farina2014extraction}; and although convolutional front-ends before a recurrent decoder are common \cite{xia2018emg,bao2021cnnlstm}, the marginal value of the convolutional stage \cite{bai2018tcn} is rarely isolated with a control that differs only in that stage.

A second question dominates deployment cost and is rarely posed with the same rigor: \emph{whose data should train the decoder for a new user?} The user's recording of the target activity is scarce; the obvious remedies are to borrow the user's recordings of \emph{other} activities, or other users' recordings of the \emph{same} activity. Both are usually evaluated by comparing ``less data'' with ``more data,'' which confounds two things. Adding data adds optimizer updates, so a gain may reflect training budget rather than new information (repeated epochs over fixed data recover much of the value of new data \cite{muennighoff2023scaling}); and the farther the source, the more sample size and distribution shift grow together, so an average gain may be the net of substantial positive and negative transfer \cite{rosenstein2005transfer,wang2019characterizing,zhang2023negative}, where which source helps which target is a directed relation rather than a similarity \cite{zamir2018taskonomy}. Transfer studies in myoelectric control, largely on gesture classification \cite{coteallard2019deep}, resolve neither confound for continuous multi-joint regression, and the conditional value of other subjects' data, its interaction with the user's own calibration budget, and its per-subject safety have not, to our knowledge, been quantified against budget-matched within-subject controls.

This letter addresses the decoder and the data question together in a staged, deliberately conservative design (Fig.~\ref{fig:design}) on all \nCohortw{} participants of a public high-density sEMG locomotion dataset \cite{dimitrov2023dataset}, with one \nChan-channel set fixed by a pre-specified channel-integrity audit. Its contributions are (i)~a compact, strictly causal decoder, \rctcn{} (\vRcParams{} parameters), selected under a held-out-trial protocol by three pre-specified contrasts---time-domain features carry the decodable information ($\dR=\vTdDRThree$ over frequency-domain features), whereas a residual causal TCN front-end ($\vTcnDRThree$) and IMU fusion ($\vImuDRThree$) are neutral on average, null results we report rather than an architectural advantage---and then frozen as the measurement instrument; (ii)~an \emph{update-matched replay control} that separates the \emph{information gain} of an added source from its \emph{budget effect}, applied to a user's own other activities (total gain $\vTotDRThree$, of which $\vInfDRThree$ is information; stair descent a reliably harmful donor, sit--stand--walk a reliably beneficial one); (iii)~the \emph{state-conditional} value of other users' same-activity data, $\GT{}=\vGtDRThree$ when the user has only the target activity and $\GW{}=\vGwDRThree$ once own other-activity data exist, with a recurring \emph{target receptivity} across sources ($r=\vRecGtRTwo$) that leave-one-participant-out (LOPO) gates identify only partly; and (iv)~a full-calibration program in which donor value collapses after one to two target records. The decoder is evaluated offline; no exoskeleton, controller, or assistance outcome is claimed.

\section{DATA, DECODER, AND PROTOCOL}
\label{sec:setup}

\subsection{Dataset, Channel Audit, and Sources}
We use the open lower-limb locomotion dataset of Dimitrov \emph{et al.} \cite{dimitrov2023dataset}: ten healthy adults (five female) with a 64-channel high-density sEMG grid below the right knee and a six-axis IMU on the tibia (both 2000\,Hz) and 100\,Hz motion capture; ground-truth hip, knee, and ankle angles are the dataset's OpenSim inverse-kinematics solutions. From the force-plate-segmented right-foot gait cycles we take six \emph{activities}---slow walking, fast walking, stair ascent, stair descent, sit--stand--walk, and stand-to-walk---with three cycles (\emph{records}, 0.9--3.7\,s) per activity and participant; a \emph{target} is one participant$\times$activity combination. sEMG was band-pass (20--450\,Hz) and notch (50\,Hz) filtered and common-average referenced; IMU channels were low-pass filtered at 15\,Hz. An audit of channels constant over an entire record (a detached or saturated electrode) found one to four dead channels in four participants; because the transfer experiments pool records across activities and participants, a dead channel would propagate into every model that borrows from an affected record, so the \nChanExclw{} channels dead in any participant (\chanExclList) are removed from every record and the remaining \nChan{} channels are used unchanged in Sections~\ref{sec:selection}--\ref{sec:stage3}, keeping all \nCohortw{} participants (\nCohortRecords{} records, \nCohortTargets{} targets). The full-calibration program of Section~\ref{sec:full} predates the audit and uses the \nFullw{} complete-grid participants with all 64 channels. The data were collected by the dataset's authors under approval of the Imperial College London Research Ethics Committee (reference 21IC6941), in accordance with the Declaration of Helsinki and with written informed consent, and released without identifiers under a CC BY 4.0 license (figshare, doi:10.6084/m9.figshare.22227337.v2) \cite{dimitrov2023dataset}; this study uses only these public, de-identified data and involved no new data collection.

\begin{figure*}[!t]
\centering
\begin{tikzpicture}[x=1cm,y=1cm,
  blk/.style={draw=#1, fill=#1!12, rounded corners=2pt, thick, align=center, inner sep=2.5pt, font=\fontsize{6.6}{7.6}\selectfont, text width=2.9cm},
  blk/.default=black,
  arr/.style={-{Latex[length=1.5mm]}, thick, draw=black!70},
  note/.style={font=\fontsize{6.2}{7.2}\selectfont, align=left, text=black!70},
  ttl/.style={font=\fontsize{6.8}{7.8}\selectfont\bfseries, anchor=west}]
\node[ttl] at (0,3.05) {(a) Source hierarchy for a target (user $u$, activity $a$)};
\node[blk=cOwn] (own) at (1.65,2.3) {\textbf{Own, same activity}\\ $u$'s other trials of $a$ (two records)};
\node[blk=cOther, right=0.3cm of own] (oth) {\textbf{Own, other activities}\\ $u$'s five other activities (15 records)};
\node[blk=cCross, right=0.3cm of oth] (xs) {\textbf{Other users, same activity}\\ the other \nCohortDonors{} participants' records of $a$};
\draw[arr] (own) -- (oth); \draw[arr] (oth) -- (xs);
\node[note, anchor=west, text width=6.2cm] at ($(xs.east)+(0.25,0)$) {$\leftarrow$ increasing source distance\\ held-out unit: one complete trial of $(u,a)$, never used for features, normalization, model selection, or gates};
\node[ttl] at (0,1.2) {(b) Update-matched replay control (Section~\ref{sec:protocol})};
\node[blk=cGray, text width=2.2cm] (t0) at (1.3,0.3) {\textbf{Target-only}\\ own same-activity data\\ $B$ optimizer updates};
\node[blk=cGray, text width=2.2cm, right=0.25cm of t0] (rp) {\textbf{Replay}\\ same data, repeated\\ $B'$ updates ($=$ source run)};
\node[blk=cOther, text width=2.2cm, right=0.25cm of rp] (src) {\textbf{+ Source}\\ own $\cup$ source data\\ $B'$ updates};
\node[note, anchor=north west, text width=8.6cm] at ($(t0.south west)+(0,-0.12)$) {total gain $=$ (+Source) $-$ (Target-only); \; information gain $=$ (+Source) $-$ (Replay);\\ budget effect $=$ (Replay) $-$ (Target-only)};
\node[ttl] at (8.9,1.2) {(c) Own-data states for other users' data (Section~\ref{sec:stage3})};
\node[blk=cCross, text width=2.2cm] (pre) at (10.2,0.3) {\textbf{Phase 1}\\ other users' same activity \emph{or} replay};
\node[blk=cOther, text width=2.2cm, right=0.25cm of pre] (mid) {\textbf{Phase 2 (W only)}\\ own other activities;\\ skipped in state T};
\node[blk=cOwn, text width=2.2cm, right=0.25cm of mid] (ft) {\textbf{Phase 3}\\ target-activity fine-tuning};
\draw[arr] (pre) -- (mid); \draw[arr] (mid) -- (ft);
\node[note, anchor=north west, text width=7.6cm] at ($(pre.south west)+(0,-0.12)$) {$\GT{}=$ cross-subject $-$ replay in T; \; $\GW{}=$ cross-subject $-$ replay in W;\\ $\Istate{}=\GW{}-\GT{}$ (displacement of others' data by own data)};
\end{tikzpicture}
\caption{Study design. (a)~Data sources for a new user $u$ performing activity $a$, ordered by distance from the target; one complete trial of $(u,a)$ is always held out. (b)~Every added source is compared against an \emph{update-matched replay} of the target's own data, which separates information from optimizer updates. (c)~Other users' same-activity data enter as a pretraining phase in two own-data states, T (target activity only) and W (plus the user's other activities). Participants, channel set, and the frozen \rctcn{} are the same throughout; only the training data change.}
\label{fig:design}
\end{figure*}
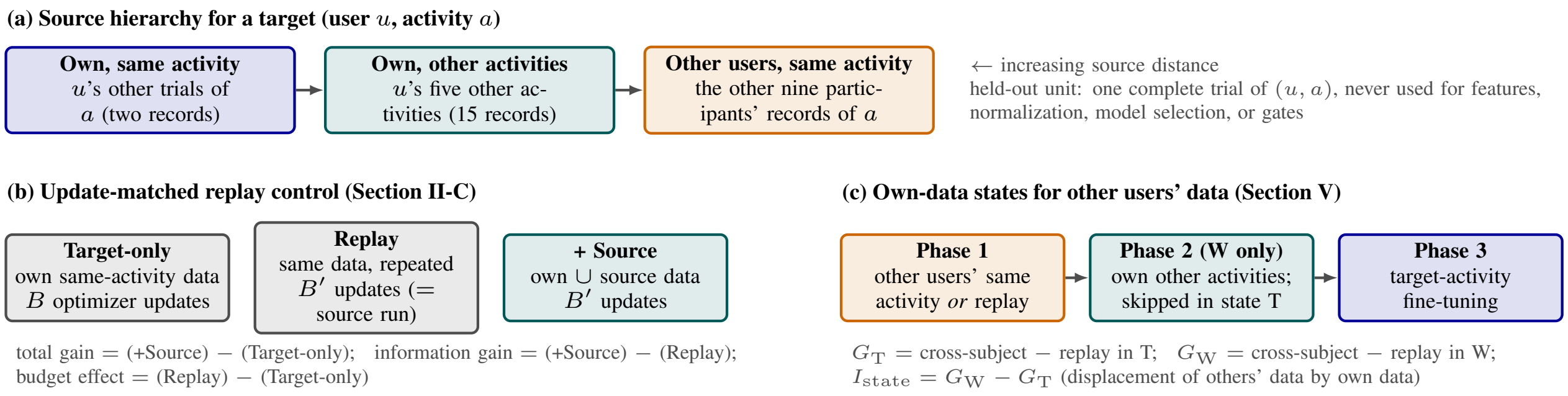

For a new user $u$ and target activity $a$, one complete trial of $(u,a)$ is always held out and never used for features, normalization, model selection, or gates; the decoder may be trained on the user's \emph{own same-activity} data (the two remaining records, the \emph{target-only} baseline), the user's \emph{own other-activity} data (15 records of the five other activities), and \emph{other users' same-activity} data (the other \nCohortDonors{} participants' records of $a$), ordered by distance from the target (Fig.~\ref{fig:design}a). For each external source we ask how much it adds, how much of that is information rather than budget, and how its value depends on what the user has already contributed.

\subsection{Frozen Decoder: \rctcn{} and a Matched Control}
\label{sec:decoder}
Raw signals are reduced to causal frame-level features at 100\,Hz: per sEMG channel, root-mean-square, mean absolute value, and waveform length per 20-sample frame, smoothed by a causal 100\,ms rolling mean (\nChanFeatThree{} features); per IMU axis, the instantaneous value, 100\,ms rolling mean and standard deviation, and first difference (24). The decoder consumes a causal window of 50 frames (0.5\,s) and, after a warm-up of $\approx$0.58\,s per record, emits predictions at 100\,Hz. Features are z-scored per participant on the fold's training records; angles and angular velocities are z-scored per joint on the target's two training records, the same constants being applied to every record of the fold's training set. A frequency-domain (FD) control replaces the time-domain (TD) EMG vector by the \nChanFeat-d vector of per-channel median frequencies of Welch and FFT spectra over the trailing 100\,ms.

\rctcn{} integrates the two encoder variants of a previously developed six-member within-subject ensemble into one strictly causal network: the EMG and IMU vectors are fused early into a \nFrameFeat-d frame vector, projected to 64 dimensions (ReLU, dropout 0.25), and passed through three residual causal convolution blocks \cite{bai2018tcn} (Conv1d $64\to64$, kernel 3, dilations 1, 2, 4, ReLU, dropout, identity skip; causal padding, receptive field 150\,ms). An 8-d learned activity embedding is concatenated at every frame, a single-layer GRU (hidden 64) supplies memory over the window, and three angle heads plus three auxiliary angular-velocity heads are trained with a Smooth-L1 angle loss ($\beta=0.5$) plus $0.2\times$ the velocity loss; only the angle heads are used at inference. The network has \vRcParams{} parameters; the \emph{Matched GRU} control removes only the three residual blocks (\vGruParams{} parameters) and isolates the residual causal TCN under identical inputs, outputs, and training. Training windows are the 50-frame sequences ending at every frame of the training records, drawn in shuffled mini-batches of 32; a run makes 18 passes ($\lfloor 18\,n_w/32\rfloor$ updates for $n_w$ windows) with AdamW (learning rate $1.2\times10^{-3}$, weight decay $2\times10^{-4}$) and fixed seeds that determine initialization, dropout masks, and batch order. All hyperparameters were frozen before any data-value experiment; the program ran on a dependency-free NumPy re-implementation of the decoders checked against the reference implementation.

\subsection{Held-Out-Trial Protocol, Replay Control, and Inference}
\label{sec:protocol}
For every target, each of its three records is held out in turn and the decoder is trained from the same initialization on the two remaining records plus the arm's source data, if any. Each held-out record yields one \R{} and one RMSE per joint; a participant's score averages its \recordsPerParticipantHeld{} held-out record$\times$joint values, the cohort score is the mean of participant scores, and every evaluation is the mean over the fixed seeds. Adding a source changes both the information available and the number of updates a fixed-epoch schedule performs; the replay control (Fig.~\ref{fig:design}b) separates them. \emph{Target-only} trains on the two target records for $B$ updates, \emph{replay} on the same records repeated until the update count equals that of the source run, $B'$, and \emph{source} on the target records together with the source data for $B'$ updates. The \emph{total gain} is source minus target-only, the \emph{information gain} source minus replay, and their difference the \emph{budget effect}. Replay matches updates, not every property of the training distribution; it is the counterfactual a practitioner can always realize, and all contrasts in Sections~\ref{sec:stage2} and \ref{sec:stage3} are relative to it unless stated.

The unit of inference is the participant \cite{lazic2010pseudoreplication}. Every effect is computed per target (pooled over the three held-out trials and, unless stated, the three joints), averaged to a participant-level effect, and summarized as the mean over participants with a participant-clustered percentile-bootstrap 95\% CI (resampling participants with replacement, keeping all of a participant's targets together), read together with the number of participants whose effect has the sign of the mean. Reported $p$-values are exact two-sided sign-flip tests over all \nSignFlips{} sign assignments of the \nPw{} participant effects with Holm's step-down correction \cite{holm1979simple} within pre-specified families; the smallest attainable $p$ is $\nMinPfrac=\nMinP$, so $p$-values are descriptive and the CIs, sign counts, and pre-registered equivalence bounds ($\pm0.01$ in \R{}, $\pm0.25$\dg{} in RMSE; an effect is practically equivalent to zero when its 95\% CI lies inside them) carry the inference. Leakage control follows one rule: nothing that touches the held-out record may influence training; normalization, features, model selection, thresholds, and compatibility estimates are fit on the fold's training records only, and every selection policy is fit inside a \nFolds-fold leave-one-participant-out (LOPO) loop whose outer participant contributes nothing to it. \emph{Oracle} policies, which select on the held-out outcomes, are non-deployable upper bounds.

\section{DECODER SELECTION UNDER THE HELD-OUT-TRIAL PROTOCOL}
\label{sec:selection}
Six configurations were trained in the target-only arm (three seeds, \nRecEval{} record$\times$joint evaluations each): \rctcn{} and the Matched GRU with sEMG+IMU or sEMG-only TD input, \rctcn{} with FD EMG input, and \rctcn{} with IMU-only input. Three contrasts were pre-specified---the residual causal TCN (\rctcn{} vs.\ Matched GRU, same input), IMU fusion (sEMG+IMU vs.\ sEMG within \rctcn{}), and the feature domain (TD vs.\ FD within \rctcn{})---and the rule to adopt the configuration that wins all three was fixed before the data-value experiments, with \rctcn{} (sEMG+IMU, TD) frozen a priori as the measurement instrument.

The three TD configurations with sEMG input are close to one another ($\R$ from \vGruRtwoThree{} for the Matched GRU with sEMG+IMU to \vGruEmgRtwo{} for the Matched GRU with sEMG only; \rctcn{} with sEMG+IMU: $\R=\vRcRtwo$, RMSE $\vRcRmse$\dg, MAE $\vRcMae$\dg, $r=\vRcR$) and far ahead of the FD ($\R=\vFdRtwo$) and IMU-only ($\vImuOnlyRtwo$) configurations. Of the three contrasts, only the feature domain is established: TD features add $\dR=\vTdDR$ ($\vTdDRCi$; $\dRMSE=\vTdDRmseTwo$\dg) in \vTdImpRw{} participants. The residual causal TCN adds nothing over the Matched GRU ($\dR=\vTcnDR$, $\vTcnDRCi$; participant effects \vTcnMin{} to \vTcnMax), and tripling the training budget on six targets did not change the sign; IMU fusion is neutral on average but strongly heterogeneous ($\dR=\vImuDR$, $\vImuDRCi$; participant effects from \vImuMin{} to \vImuMax). The selection rule is therefore not decided by the data; we retained \rctcn{}, frozen before this analysis (the Matched GRU would have served equally well). Absolute levels are lower than at full calibration (Section~\ref{sec:full}) because each network sees only two records of one activity; with architecture, seeds, optimizer, and schedule fixed, every effect in Sections~\ref{sec:stage2} and \ref{sec:stage3} is attributable to the composition of the training set.

\section{THE VALUE OF A USER'S OWN OTHER-ACTIVITY DATA}
\label{sec:stage2}
For every target and held-out trial, three networks are trained from the same initialization: \emph{target-only} (the two remaining records of $(u,a)$), \emph{replay} (the same two records, update-matched to the pooled run), and \emph{pooled} (the two target records together with the participant's 15 records of the five other activities, activity-conditioned through the embedding).

\begin{table}[!t]
\caption{Stage 2: Data-State Levels and Primary Effects (\nPw{} Participants, \nTargets{} Targets, Held-Out Trials; Participant-Clustered 95\% CIs)}
\label{tab:stage2}
\centering
\scriptsize
\setlength{\tabcolsep}{2.5pt}
\begin{threeparttable}
\begin{tabular}{@{}lccccc@{}}
\toprule
State & \R{} & RMSE (\dg) & MAE (\dg) & $r$ & CCC \\
\midrule
Target-only & 0.7222 & 6.681 & 5.146 & 0.896 & 0.853 \\
Replay (update-matched) & 0.7400 & 6.368 & 4.910 & 0.903 & 0.858 \\
Pooled (+ own other activities) & 0.7880 & 5.631 & 4.399 & 0.929 & 0.893 \\
\bottomrule
\end{tabular}
\vspace{3pt}
\begin{tabular}{@{}llcc@{}}
\toprule
Effect & Outcome & Mean [95\% CI] & Impr. \\
\midrule
Total gain & \dR{} & $+0.0658$ $[+0.0361,\,+0.0955]$ & 10/10 \\
\quad(pooled $-$ target-only) & \dRMSE{} & $-1.049$\dg{} $[-1.462,\,-0.649]$ & 10/10 \\
Information gain & \dR{} & $+0.0480$ $[+0.0166,\,+0.0790]$ & 7/10 \\
\quad(pooled $-$ replay) & \dRMSE{} & $-0.736$\dg{} $[-1.122,\,-0.339]$ & 9/10 \\
Budget effect\tnote{a} & \dR{} & $+0.0178$ & -- \\
\quad(replay $-$ target-only) & \dRMSE{} & $-0.313$\dg{} & -- \\
\bottomrule
\end{tabular}
\begin{tablenotes}
\item[a] Difference of cohort means; not a pre-registered estimand, so no CI is reported.
\end{tablenotes}
\end{threeparttable}
\end{table}

\begin{figure*}[!t]
\centering
\includegraphics[width=0.84\textwidth]{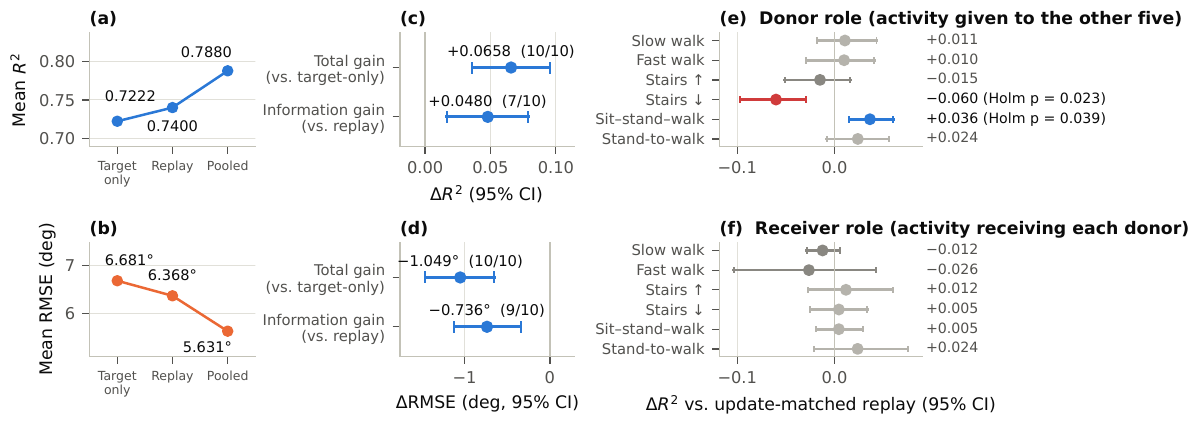}
\caption{The value of a user's own other-activity data. (a,\,b)~Cohort-mean \R{} and RMSE of the three data states. (c,\,d)~Total gain and information gain with 95\% participant-clustered CIs (participants improved in parentheses); blue marks CIs that exclude zero. (e)~Donor role: mean \dR{} when each activity is the single donor to the other five targets, relative to replay; stair descent (harmful) and sit--stand--walk (beneficial) are the Holm-significant roles. (f)~Receiver role: mean \dR{} of each target over its five single donors; none survives Holm correction.}
\label{fig:stage2}
\end{figure*}

\emph{The total gain is real, and most of it is information.} Adding a user's own other-activity data raises the cohort mean from $\R=\vSToRtwo$ to $\vSPoRtwo$ and lowers RMSE from \vSToRmse\dg{} to \vSPoRmse\dg{} (Table~\ref{tab:stage2}, Fig.~\ref{fig:stage2}a--d): the total gain is $\dR=\vTotDR$ ($\vTotDRCi$; \vTotImpRw{} participants) and $\dRMSE=\vTotDRmse$\dg{} ($\vTotDRmseCi$). Replay alone reaches $\R=\vSRpRtwo$, so the budget effect is $\dR=\vBudDR$ and $\dRMSE=\vBudDRmse$\dg---\vBudShareR{} of the \R{} gain and \vBudShareRmse{} of the RMSE gain that a naive ``more data'' comparison would attribute to the other activities are obtained by training longer on the target data alone---and the information gain beyond replay is $\dR=\vInfDR$ ($\vInfDRCi$; \vInfImpRw{} participants) and $\dRMSE=\vInfDRmse$\dg{} ($\vInfDRmseCi$; \vInfImpRmsew), both with CIs that exclude zero. The gain is not universal: \vInfNonImpN{} participants (\vInfNonImpList) show no information gain in \R{}.

\emph{Transfer is directional and role-asymmetric.} Each of the \vDirectedPairs{} directed donor$\to$target activity pairs was evaluated by adding the three records of a single donor activity to the target's two records against its own update-matched replay, and the pairs were aggregated into six \emph{donor roles} (an activity's mean effect on the other five targets) and six \emph{receiver roles}. None of the \vDirectedPairs{} pairs survives Holm correction over the \vDirectedTests{} pair-level tests---the data do not support a pairwise transfer map---but the role aggregates are informative (Fig.~\ref{fig:stage2}e,\,f). Stair descent as a donor is reliably harmful: $\dR=\vSdDonorDR$ ($\vSdDonorDRCi$), $\dRMSE=\vSdDonorDRmse$\dg{} ($\vSdDonorDRmseCi$), Holm $p=\vSdDonorPholm$ over the six donor roles, the worst participant losing $\vSdDonorWorstR$ in \R{}. Sit--stand--walk as a donor is reliably beneficial: $\dR=\vSsDonorDR$ ($\vSsDonorDRCi$; Holm $p=\vSsDonorPholm$). Every other donor role lies within $\pm\vOtherDonorMaxAbs$; no receiver role survives correction: level walking tends to be harmed by a single foreign activity (fast walking $\dR=\vRecvFwDR$, slow walking $\vRecvSwDR$), stand-to-walk ($\vRecvStwDR$) and stair descent ($\vRecvSdDR$) tend to benefit. Two features matter for practice: the effect of a source depends on the direction of transfer (stair descent is a harmful donor but a neutral-to-receptive target), and single-donor increments are on average null (mean receiver-role \dR{} $\vRecvMeanDR$) while the five-donor pool gains $\vInfDRThree$---mixing activities averages out the biases any one foreign activity imposes.

\emph{Timescale mismatch is associated with worse transfer.} Among ten \emph{training-visible descriptors} computed from the fold's training records only---three of the target's own-data sufficiency (number of windows, record duration, excursion), one of difficulty (cross-record linear decodability), and six of source--target compatibility (duration mismatch, prototype-trajectory correlation and RMSE, EMG and IMU feature-distribution shift, linear transferability)---only the standardized mismatch in record duration is consistently associated with the pair-level effect: $\dR=\vDurDR$ per SD ($\vDurDRCi$; clustered sign-flip $p=\vDurDRP$), $\dRMSE=\vDurDRmse$\dg{} ($\vDurDRmseCi$; Holm $p=\vDurPholm$ over the two outcomes), sign preserved under leave-one-participant-out. It is collinear with the other descriptors (VIF \vDurVif) and adds no out-of-participant predictive value: an association, not a rule (Section~\ref{sec:policy}).

\section{THE VALUE OF OTHER USERS' DATA, CONDITIONAL ON THE USER'S OWN DATA}
\label{sec:stage3}
Other users' same-activity data enter as a pretraining phase (Fig.~\ref{fig:design}c). Phase~1 trains the frozen decoder either on the other \nDonors{} participants' records of the target activity (\emph{cross-subject}) or on an update-matched replay of the target's two records; phase~3 fine-tunes on the target's two records. State~T (target activity only) goes directly from phase~1 to phase~3; state~W inserts a phase~2 on the participant's pooled Stage-2 training set (the two target records plus its 15 other-activity records). No participant embedding, adapter, or branch is added; the network is identical in all four arms. The estimands are $\GT{}=\mathrm{score}(\text{cross-subject}\mid\mathrm{T})-\mathrm{score}(\text{replay}\mid\mathrm{T})$, \GW{} likewise in state W, and $\Istate{}=\GW{}-\GT{}$, the displacement of others' data by own data.

\begin{table}[!t]
\caption{Stage 3: Data-State Levels and Primary Estimands (\nPw{} Participants, \nTargets{} Targets; Participant-Clustered 95\% CIs; Exact Sign-Flip $p$, Holm-Corrected Within Three Estimands per Outcome; Equiv.\ = CI Inside the Pre-Registered Equivalence Bound)}
\label{tab:stage3}
\centering
\scriptsize
\setlength{\tabcolsep}{2pt}
\begin{tabular}{@{}lccccc@{}}
\toprule
State & Phase-1 source & \R{} & RMSE (\dg) & MAE (\dg) & $r$ \\
\midrule
T & replay & 0.7444 & 6.323 & 4.915 & 0.905 \\
T & cross-subject & 0.7969 & 5.452 & 4.287 & 0.931 \\
W & replay & 0.7823 & 5.605 & 4.394 & 0.926 \\
W & cross-subject & 0.8100 & 5.220 & 4.096 & 0.933 \\
\midrule
no target data & cross-subject only & 0.2098 & 10.764 & 8.797 & 0.805 \\
\bottomrule
\end{tabular}
\vspace{3pt}
\begin{tabular}{@{}llcccc@{}}
\toprule
Estimand & Outcome & Mean [95\% CI] & Impr. & $p_{\mathrm{Holm}}$ & Equiv. \\
\midrule
\GT{} & \dR{} & $+0.0525$ $[+0.0189,\,+0.0872]$ & 8/10 & $0.070$ & no \\
 & \dRMSE{} & $-0.871$\dg{} $[-1.292,\,-0.455]$ & 9/10 & $0.018$ & no \\
\GW{} & \dR{} & $+0.0277$ $[+0.0072,\,+0.0486]$ & 7/10 & $0.098$ & no \\
 & \dRMSE{} & $-0.385$\dg{} $[-0.643,\,-0.152]$ & 9/10 & $0.018$ & no \\
\Istate{} & \dR{} & $-0.0249$ $[-0.0542,\,+0.0005]$ & 3/10 & $0.11$ & no \\
 & \dRMSE{} & $+0.486$\dg{} $[+0.224,\,+0.758]$ & 1/10 & $0.018$ & no \\
\bottomrule
\end{tabular}
\end{table}

\begin{figure*}[!t]
\centering
\includegraphics[width=0.84\textwidth]{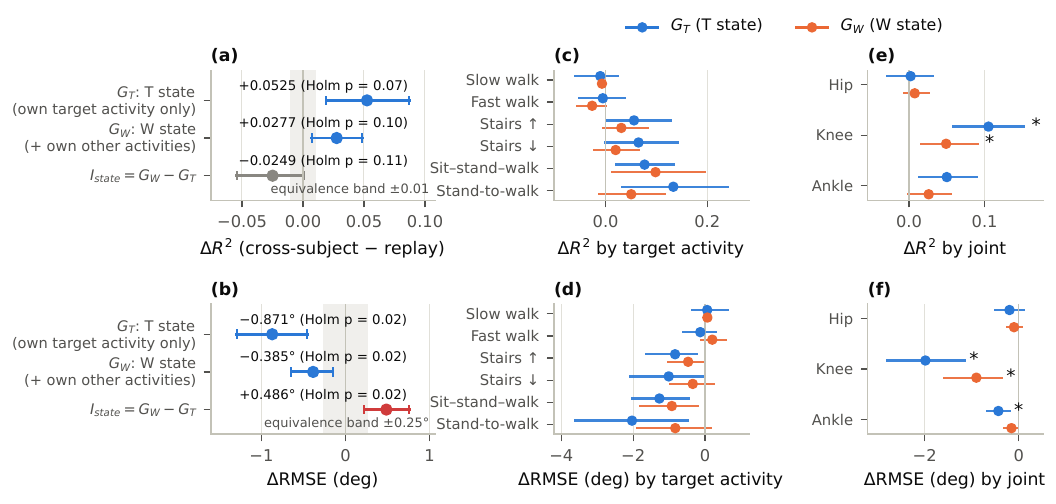}
\caption{The value of other users' same-activity data depends on the user's own-data state. (a,\,b)~The three estimands with 95\% participant-clustered CIs and the pre-registered equivalence bands; gray marks CIs that include zero, red a CI that excludes zero in the harmful direction. (c,\,d)~\GT{} (blue) and \GW{} (orange) by target activity and (e,\,f)~by joint; asterisks mark Holm-significant joint-level effects.}
\label{fig:stage3}
\end{figure*}

\emph{Other users' data help at low calibration, and less once own data are available.} When the user has contributed only the target activity, pretraining on other users is beneficial (Table~\ref{tab:stage3}, Fig.~\ref{fig:stage3}a,\,b): $\GT{}=\vGtDR$ ($\vGtDRCi$; \vGtImpRw{} participants) and $\dRMSE=\vGtDRmse$\dg{} ($\vGtDRmseCi$), surviving Holm correction in RMSE ($p_{\mathrm{Holm}}=\vGtPholmRmse$) and just missing it in \R{} ($\vGtPholmR$). Other users' data are a prior, not a substitute: without the target's records the cross-subject network reaches only $\R=\vSPcRtwoThree$. Once the user's other activities are present, a smaller residual value remains, $\GW{}=\vGwDR$ ($\vGwDRCi$) and $\dRMSE=\vGwDRmse$\dg{} ($\vGwDRmseCi$; $p_{\mathrm{Holm}}=\vGwPholmRmse$), so neither estimand is practically equivalent to zero; and own data displace others' data, $\Istate{}=\vIsDR$ in \R{} ($\vIsDRCi$) and $\vIsDRmse$\dg{} in RMSE ($\vIsDRmseCi$; $p_{\mathrm{Holm}}=\vIsPholmRmse$), with \vIsNotImpRmsew{} participants losing cross-subject value between T and W. The best state overall is \vBestStateName{} ($\R=\vSWcRtwo$, RMSE \vSWcRmse\dg)---other users' same activity, then the user's own other activities, then target fine-tuning. The cohort means hide heterogeneity (Fig.~\ref{fig:stage3}c--f): in state T the transitions receive other users' data best (stand-to-walk $\GT{}=\vGtStwDR$, sit--stand--walk $\vGtSsDR$) and level walking gains nothing (fast walking $\vGtFwDR$, slow walking $\vGtSwDR$), although no activity-level test survives correction; across joints the knee carries the benefit ($\GT{}=\vGtKneeDR$; Holm $p=\vGtKneePholm$) and the hip gains nothing ($\vGtHipDR$).

\begin{figure}[!t]
\centering
\includegraphics[width=0.85\columnwidth]{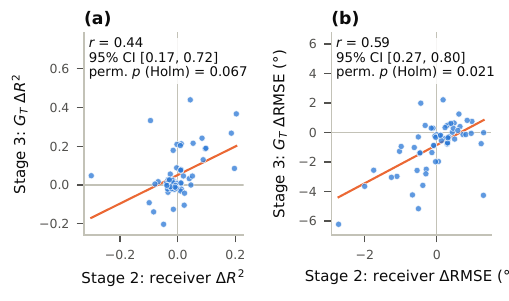}
\caption{Target receptivity recurs across data sources. Each point is one of the \nTargets{} targets: its Stage-2 receiver-role effect (mean single-donor increment relative to replay) against its Stage-3 \GT{}, in (a)~\dR{} and (b)~\dRMSE{}; least-squares fits with $r$, its 95\% bootstrap CI, and the Holm-corrected activity-label permutation $p$.}
\label{fig:receptivity}
\end{figure}

\emph{Target receptivity recurs across sources.} For each of the \nTargets{} targets, the Stage-2 receiver-role effect and the Stage-3 \GT{} are positively correlated (Fig.~\ref{fig:receptivity}): $r=\vRecGtR$ ($\vRecGtRCi$) in \dR{} and $\vRecGtRmse$ ($\vRecGtRmseCi$) in \dRMSE{}; activity-label permutation tests give Holm-corrected $p=\vRecGtRPholm$ and $\vRecGtRmsePholm$, and the correlation with \GW{} is weaker ($r=\vRecGwRTwo$, $\vRecGwRmseTwo$), as expected once own data have displaced part of the cross-subject value. A target that absorbs its user's other activities tends to absorb other users' same activity as well; because both effects are measured on the same held-out records, part of the correlation may be record-level idiosyncrasy, and receptivity is a phenomenon to be explained, not an explanation.

\begin{table*}[!t]
\caption{Deployable Selection Policies Versus Non-Deployable Oracles, Relative to Update-Matched Replay (\nPw{} Participants, \nTargets{} Targets; Participant-Clustered 95\% CIs)}
\label{tab:policies}
\centering
\scriptsize
\setlength{\tabcolsep}{2pt}
\begin{threeparttable}
\begin{tabular}{@{}llccccc@{}}
\toprule
Regime & Policy & Selected & \dR{} [95\% CI] & \dRMSE{} (\dg) [95\% CI] & Pos.\,/\,neg.\tnote{a} & Worst participant \dR{} \\
\midrule
\multirow{4}{*}{\makecell[l]{Own other\\activities\\(Stage 2)}} & Always pool the five other activities & 60/60 & $+0.0480$ $[+0.0166,\,+0.0790]$ & $-0.736$ $[-1.122,\,-0.339]$ & --\,/\,24 & $-0.021$ \\
 & Fixed single donor (chosen on training participants) & 60/60 & $+0.0408$ $[+0.0121,\,+0.0755]$ & $-0.488$ $[-0.876,\,-0.134]$ & --\,/\,23 & $-0.028$ \\
 & LOPO compatibility safe gate & 20/60 & $+0.0024$ $[-0.0228,\,+0.0195]$ & $-0.048$ $[-0.204,\,+0.167]$ & 11\,/\,9 & $-0.099$ \\
 & Oracle: best positive single donor\tnote{b} & 53/60 & $+0.0809$ $[+0.0534,\,+0.1124]$ & $-0.991$ $[-1.298,\,-0.706]$ & 53\,/\,0 & $+0.025$ \\
\midrule
\multirow{3}{*}{\makecell[l]{Other users,\\state T\\(Stage 3)}} & Always cross-subject pretraining & 60/60 & $+0.0525$ $[+0.0189,\,+0.0872]$ & $-0.871$ $[-1.292,\,-0.455]$ & 40\,/\,20 & $-0.026$ \\
 & LOPO safe gate & 28/60 & $+0.0362$ $[+0.0115,\,+0.0618]$ & $-0.590$ $[-0.908,\,-0.293]$ & 20\,/\,8 & $-0.035$ \\
 & Oracle: cross-subject where positive\tnote{b} & 40/60 & $+0.0675$ $[+0.0400,\,+0.0959]$ & $-1.048$ $[-1.418,\,-0.671]$ & 40\,/\,0 & $+0.004$ \\
\midrule
\multirow{3}{*}{\makecell[l]{Other users,\\state W\\(Stage 3)}} & Always cross-subject pretraining & 60/60 & $+0.0277$ $[+0.0072,\,+0.0486]$ & $-0.385$ $[-0.643,\,-0.152]$ & 35\,/\,25 & $-0.017$ \\
 & LOPO safe gate & 17/60 & $+0.0086$ $[-0.0048,\,+0.0250]$ & $-0.079$ $[-0.260,\,+0.072]$ & 11\,/\,6 & $-0.016$ \\
 & Oracle: cross-subject where positive\tnote{b} & 35/60 & $+0.0419$ $[+0.0245,\,+0.0605]$ & $-0.536$ $[-0.765,\,-0.331]$ & 35\,/\,0 & $+0.005$ \\
\bottomrule
\end{tabular}
\begin{tablenotes}
\item[a] Selected targets whose held-out \dR{} is positive\,/\,negative; for the Stage-2 ``always'' policies, only the number worse than replay is reported (\vPtwoPoolNegPctText{} and \vPtwoFixedNegPctText{} of \nTargets).
\item[b] Oracles select using the held-out outcomes and are upper bounds, not deployable policies.
\end{tablenotes}
\end{threeparttable}
\end{table*}

\subsection{Predictability: LOPO Gates Versus the Oracle}
\label{sec:policy}
Receptivity would be actionable if it could be predicted before the held-out record is seen. LOPO predictors of the per-target Stage-3 effect from the ten descriptors (ridge regressions on growing descriptor subsets and a random forest, tuned by an inner LOPO loop) did not improve on the training participants' mean effect for \GT{} (Holm $p\ge\vLopoGtPholmMin$), and the richest model was significantly worse for \GW{} (Holm $p=\vLopoGwWorseP$); the descriptors are collinear (VIF up to \vMaxVif). Table~\ref{tab:policies} gives the deployment consequence, comparing on identical held-out records and relative to replay \emph{always transferring}, a \emph{fixed single donor} chosen on each fold's training participants, a \emph{LOPO safe gate} (an $\ell_2$-logistic classifier of positive transfer on the ten descriptors, fitted inside the LOPO loop, transferring at predicted probability $\ge0.7$), and the \emph{oracle} that transfers exactly where the held-out effect is positive. For the user's own other activities, always pooling ($\vPtwoPoolDR$; \vPtwoPoolNegPct{} of targets worse than replay) and a fixed single donor ($\vPtwoFixedDR$; \vPtwoFixedNegPct{}) beat the safe gate, which selects \vPtwoGateSelN{} of \nTargets{} targets and neither helps nor hurts ($\vPtwoGateDR$, $\vPtwoGateDRCi$); the oracle would reach $\vPtwoOracleDR$ ($\vPtwoOracleDRCi$) with every participant improving, so a beneficial single donor exists for \vPtwoOracleSelN{} targets but the descriptors do not identify it, and the safest deployable choice is to pool all other activities. For other users' data in state T, always pretraining gains $\vPthreeTAlwaysDR$ but leaves \vPthreeTAlwaysNeg{} of \nTargets{} targets worse than replay; the safe gate selects \vPthreeTGateSelN{} targets (\vPthreeTGatePos{} positive, \vPthreeTGateNeg{} negative) and attains $\dR=\vPthreeTGateDR$ ($\vPthreeTGateDRCi$) and $\dRMSE=\vPthreeTGateDRmse$\dg---two thirds of the value of always transferring with half the harmed targets, the only deployable gate whose intervals exclude zero---against an oracle of $\vPthreeTOracleDR$ ($\vPthreeTOracleDRCi$). In state W the gate gains nothing established ($\vPthreeWGateDR$, $\vPthreeWGateDRCi$) while always transferring still gains $\vPthreeWAlwaysDR$ and the oracle ceiling shrinks to $\vPthreeWOracleDR$: once own data are sufficient, less is left for any selector to find. Across both sources the ordering is oracle $>$ always-transfer $\ge$ safe gate $\ge$ replay, with \vPthreeTOracleRatio{} to \vPtwoOracleRatio{} times the average gain available to an oracle.

\section{THE FULL-CALIBRATION REGIME}
\label{sec:full}
Sections~\ref{sec:stage2} and \ref{sec:stage3} characterize the low-calibration end of the axis, where a user has contributed two records of the target activity. A separate program characterizes the other end, where a user has contributed all 17 remaining records, using the frozen six-member GRU ensemble from which \rctcn{} was derived as the within-subject reference (\nFullw{} complete-grid participants, 64 channels, each of the 18 records held out in turn; macro-average $\R=0.834$, RMSE $5.66$\dg). Participants share kinematic shape (between-participant prototype-trajectory correlation 0.777 versus 0.930 within) but not amplitude (raw-angle RMSE 11.78\dg{} versus 5.56\dg{}): someone else's angles do not predict mine frame-by-frame. Sixteen cross-subject route families, from supervised pooling and donor pretraining with fine-tuning to frozen representations, similarity-weighted mixing, cross-fitted gates, shrinkage toward donor models, donor-side self-supervision, and donor angle-structure priors, were each compared against their own matched target-only control under pre-registered gates (mean improvement, worst-subject safety, across-activity stability). Donor value exists but is small and unsafe: naive supervised pooling improves the average ($\dR=+0.042$) but only 4/6 subjects benefit and the worst declines by $-0.080$; selection, gating, and shrinkage create no information; unlabeled donor sensors carry a weak signal (donor-side self-supervision $+0.029$, all six subjects), whereas every route that injects donor \emph{angle} structure directly is null or strongly harmful ($-0.064$, $-0.075$). The one strong positive is supervised donor pretraining at low calibration: pretraining on the five donors' 90 labeled records and fine-tuning with one same-activity calibration record gives $\dR=+0.0695$ ($\dRMSE=-1.12$\dg; 5/6 subjects; subject-cluster bootstrap 95\% CI $\ci{+0.0041}{+0.1136}$ against a fully matched control), yet fails the safety gate because one subject declines by $-0.0875$, concentrated in stair ascent, the activity with the weakest between-participant correlation.

\begin{figure}[!t]
\centering
\includegraphics[width=0.7\columnwidth]{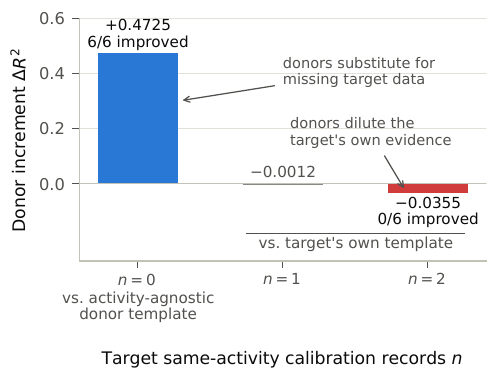}
\caption{Full-calibration regime: donor increment as a function of the target's same-activity calibration budget $n$ under protocol-time templates (\nFullw{} participants, 64 channels). At $n{=}0$ the increment is measured against an activity-agnostic donor template; at $n{=}1,2$ against the target's own template.}
\label{fig:calibration}
\end{figure}

The calibration-dependence curve (Fig.~\ref{fig:calibration}) isolates \emph{when} donor information helps: with no target same-activity record, same-activity donor structure is enormously better than an activity-agnostic donor template ($\dR=+0.4725$; 6/6 subjects); with one record, adding donors to the target's own template is already null ($-0.0012$); with two it is harmful ($-0.0355$; 0/6). Obtained with a different model family on the complete-grid subset, this curve is the full-calibration counterpart of Section~\ref{sec:stage3}: other users' data substitute for missing target-activity evidence and are diluted---then inverted---by the user's own same-activity data.

\section{DISCUSSION AND LIMITATIONS}
\label{sec:discussion}
\emph{A unified reading.} The held-out-trial and full-calibration programs agree on one picture: the marginal value of an external source decreases with the sufficiency of the user's own data, whatever the source. A user's own other activities produce a real total gain of which about three quarters survives budget matching ($\vInfDRThree$ beyond replay); other users' same-activity data are worth $\vGtDRThree$ as a pretraining prior when the user has only the target activity and half of that ($\vGwDRThree$) once own other-activity data exist; and with 17 own records no route among sixteen safely beats the within-subject ensemble, the donor increment collapsing from $+0.47$ at $n{=}0$ to null at $n{=}1$ and harmful at $n{=}2$. Two choices make this visible: the replay control, which separates information from optimization budget---a confound that, to our knowledge, has not been controlled in myoelectric transfer studies---and shows that \vBudShareR{} of what a ``more data'' comparison would call cross-activity transfer is budget; and participant-level inference with exact tests, Holm correction, and equivalence bounds.

\emph{Why average gains are small.} The averages are the net of substantial positive and negative transfer: donor roles are asymmetric, receptivity recurs across sources (the transitions receive best, level walking gains from neither), and oracles have headroom ($\vPtwoOracleDRThree$ and $\vPthreeTOracleDRThree$ against replay, every participant improving); as in the directed task-transfer graphs of other domains \cite{zamir2018taskonomy}, which source helps which target is a relation, not a similarity, and averaging over it understates what selective transfer could achieve. We have not reduced receptivity to a measurable cause: timescale mismatch is associated with worse transfer but adds no out-of-participant predictive value, and the one gate that works---the logistic gate at low calibration---ranks targets without explaining them.

\emph{Metric and benchmark context.} Hip and ankle estimates show low and often negative \R{} in low-excursion phases, where sub-degree residuals still yield $\R\ll0$; RMSE in degrees is therefore reported alongside \R{}, and the Stage-3 estimands survive correction in RMSE but not in \R{}. The RMSE levels (5.66\dg{} at full calibration, \vSToRmse\dg{} with two training records) are of the same order as within-subject lower-limb results under more permissive protocols \cite{zhang2012semg} and not comparable with results in normalized units \cite{gautam2020myonet}; here three joints are estimated jointly across six activities including transitions, with strict held-out records and participant-level uncertainty.

\emph{Implications for calibration protocols.} (1)~Record the target activities first. (2)~Extract the user's own data fully: train to a fixed update budget (the budget effect alone was worth $\vBudDRThree$ in \R{}) and pool the user's other activities with the target records ($\vInfDRThree$ beyond replay). (3)~Start from other users' same-activity data as a pretraining prior ($\GT{}=\vGtDRThree$; a conservative LOPO gate halves the \vPthreeTAlwaysNegPct{} per-target risk of doing worse than replay while keeping most of the gain), and keep it as the first phase once the user's other activities arrive: other users $\to$ own other activities $\to$ target was the best schedule of the study ($\R=\vSWcRtwoThree$). (4)~Do not curate single foreign activities as donors. (5)~At full calibration, donor-pretrained weights help only for the first minutes of use; the advantage vanishes, and risks inverting, once two same-activity records exist.

\emph{Limitations.} (i)~Ten participants bound the resolution of participant-level tests (exact $p\ge\nMinPfrac$), so Holm-corrected null results at the activity, joint, and pair level mean ``not established,'' not ``absent.'' (ii)~The full-calibration program rests on six complete-grid participants, 64 channels, and a different model family; its conclusions are re-examined, not reproduced, at low calibration. (iii)~Six activities from one dataset do not span the locomotion domain. (iv)~The Matched GRU control removes the TCN block rather than replacing it with equal recurrent capacity, so the TCN contrast bundles inductive bias with parameter count. (v)~The decoder-selection and Stage-2 state analyses use three seeds per network, the directed-pair and Stage-3 analyses \vSeedsPairStage{}, so their participant-level effects carry additional seed variance. (vi)~Descriptors are partly collinear and no mechanistic attribution is made. (vii)~The activity embedding assumes the activity label is available at inference; in deployment a locomotion-mode classifier \cite{young2014training} would have to supply it. (viii)~The decoding module is evaluated alone, offline; closed-loop use in a co-adaptive exoskeleton \cite{madduri2023biosignal} additionally requires online latency evaluation, a coupling to assistance, measured rather than assumed human adaptation, stability guarantees for the coupled two-learner loop \cite{muller2017twolearners}, and a transfer gate validated on held-out participants, of which this study provides only a low-calibration version.

\section{CONCLUSION}
With a compact causal decoder frozen as a measurement instrument, an update-matched replay control, and participant-level inference, we have shown that the value of data for continuous lower-limb intent decoding is hierarchical, state-dependent, and directional: most of a user's own other-activity value is information rather than budget ($\vInfDRThree$ beyond replay), transfer between activities has reliably harmful and reliably beneficial donors, other users' same-activity data help as a prior at low calibration ($\GT{}=\vGtDRThree$) and half as much once own other-activity data exist ($\GW{}=\vGwDRThree$), and LOPO gates recover only part of the oracle headroom. The bottleneck has moved from acquiring data to deciding, without test labels, when transfer should be allowed. The data are public; the channel audit, per-record results, and the code regenerating every statistic, table, and figure will be released.

\section*{ACKNOWLEDGMENT}
The authors thank the authors of the open dataset for releasing it. This research received no specific grant from any funding agency. A large language model (Claude, Anthropic) was used to draft and edit the text of this article under the authors' direction; the study design, analyses, results, figures, and conclusions are the authors' own, and the authors reviewed and take full responsibility for all content.

\balance
\bibliographystyle{IEEEtran}
\bibliography{references}

\begin{thebibliography}{10}
\providecommand{\url}[1]{#1}
\csname url@samestyle\endcsname
\providecommand{\newblock}{\relax}
\providecommand{\bibinfo}[2]{#2}
\providecommand{\BIBentrySTDinterwordspacing}{\spaceskip=0pt\relax}
\providecommand{\BIBentryALTinterwordstretchfactor}{4}
\providecommand{\BIBentryALTinterwordspacing}{\spaceskip=\fontdimen2\font plus
\BIBentryALTinterwordstretchfactor\fontdimen3\font minus
  \fontdimen4\font\relax}
\providecommand{\BIBforeignlanguage}[2]{{%
\expandafter\ifx\csname l@#1\endcsname\relax
\typeout{** WARNING: IEEEtranM.bst: No hyphenation pattern has been}%
\typeout{** loaded for the language `#1'. Using the pattern for}%
\typeout{** the default language instead.}%
\else
\language=\csname l@#1\endcsname
\fi
#2}}
\providecommand{\BIBdecl}{\relax}
\BIBdecl

\bibitem{collins2015reducing}
S.~H. Collins, M.~B. Wiggin, and G.~S. Sawicki, ``Reducing the energy cost of
  human walking using an unpowered exoskeleton,'' \emph{Nature}, vol. 522, pp.
  212--215, 2015.
\bibitem{slade2022personalizing}
P.~Slade, M.~J. Kochenderfer, S.~L. Delp, and S.~H. Collins, ``Personalizing
  exoskeleton assistance while walking in the real world,'' \emph{Nature}, vol.
  610, pp. 277--282, 2022.
\bibitem{molinaro2024estimating}
D.~D. Molinaro, I.~Kang, and A.~J. Young, ``Estimating human joint moments
  unifies exoskeleton control, reducing user effort,'' \emph{Sci. Robot.},
  vol.~9, no.~88, p. eadi8852, 2024.
\bibitem{gordon2007learning}
K.~E. Gordon and D.~P. Ferris, ``Learning to walk with a robotic ankle
  exoskeleton,'' \emph{J. Biomech.}, vol.~40, no.~12, pp. 2636--2644,
  2007.
\bibitem{madduri2023biosignal}
M.~M. Madduri, S.~A. Burden, and A.~L. Orsborn, ``Biosignal-based co-adaptive
  user--machine interfaces for motor control,'' \emph{Curr. Opin. Biomed.
  Eng.}, vol.~27, p. 100462, 2023.
\bibitem{zhang2012semg}
F.~Zhang, P.~Li, Z.-G. Hou, Z.~Lu, Y.~Chen, Q.~Li, and M.~Tan, ``{sEMG}-based
  continuous estimation of joint angles of human legs by using {BP} neural
  network,'' \emph{Neurocomputing}, vol.~78, no.~1, pp. 139--148, 2012.
\bibitem{xia2018emg}
P.~Xia, J.~Hu, and Y.~Peng, ``{EMG}-based estimation of limb movement using
  deep learning with recurrent convolutional neural networks,'' \emph{Artif.
  Organs}, vol.~42, no.~5, pp. E67--E77, 2018.
\bibitem{bao2021cnnlstm}
T.~Bao, S.~A.~R. Zaidi, S.~Xie, P.~Yang, and Z.-Q. Zhang, ``A {CNN-LSTM} hybrid
  model for wrist kinematics estimation using surface {EMG},'' \emph{IEEE
  Trans. Instrum. Meas.}, vol.~70, p. 2503809, 2021.
\bibitem{gautam2020myonet}
A.~Gautam, M.~Panwar, D.~Biswas, and A.~Acharyya, ``{MyoNet}: A
  transfer-learning-based {LRCN} for lower limb movement recognition and knee
  joint angle prediction for remote monitoring of rehabilitation progress from
  {sEMG},'' \emph{IEEE J. Transl. Eng. Health Med.}, vol.~8, p. 2100310,
  2020.
\bibitem{farina2014extraction}
D.~Farina \emph{et~al.}, ``The extraction of neural information from the
  surface {EMG} for the control of upper-limb prostheses: emerging avenues and
  challenges,'' \emph{IEEE Trans. Neural Syst. Rehabil. Eng.}, vol.~22, no.~4,
  pp. 797--809, 2014.
\bibitem{bai2018tcn}
S.~Bai, J.~Z. Kolter, and V.~Koltun, ``An empirical evaluation of generic
  convolutional and recurrent networks for sequence modeling,'' \emph{arXiv
  preprint arXiv:1803.01271}, 2018.
\bibitem{muennighoff2023scaling}
N.~Muennighoff \emph{et~al.}, ``Scaling data-constrained language models,'' in
  \emph{Proc. NeurIPS}, 2023.
\bibitem{rosenstein2005transfer}
M.~T. Rosenstein, Z.~Marx, L.~P. Kaelbling, and T.~G. Dietterich, ``To transfer
  or not to transfer,'' in \emph{NIPS Workshop on Inductive Transfer},
  2005.
\bibitem{wang2019characterizing}
Z.~Wang, Z.~Dai, B.~P{\'o}czos, and J.~Carbonell, ``Characterizing and avoiding
  negative transfer,'' in \emph{Proc. IEEE/CVF CVPR}, 2019, pp.
  11\,285--11\,294.
\bibitem{zhang2023negative}
W.~Zhang, L.~Deng, L.~Zhang, and D.~Wu, ``A survey on negative transfer,''
  \emph{IEEE/CAA J. Autom. Sinica}, vol.~10, no.~2, pp. 305--329, 2023.
\bibitem{zamir2018taskonomy}
A.~R. Zamir, A.~Sax, W.~Shen, L.~J. Guibas, J.~Malik, and S.~Savarese,
  ``Taskonomy: Disentangling task transfer learning,'' in \emph{Proc. IEEE/CVF
  CVPR}, 2018, pp. 3712--3722.
\bibitem{coteallard2019deep}
U.~C{\^o}t{\'e}-Allard \emph{et~al.}, ``Deep learning for electromyographic
  hand gesture signal classification using transfer learning,'' \emph{IEEE
  Trans. Neural Syst. Rehabil. Eng.}, vol.~27, no.~4, pp. 760--771, 2019.
\bibitem{dimitrov2023dataset}
H.~Dimitrov, A.~M.~J. Bull, and D.~Farina, ``High-density {EMG}, {IMU},
  kinetic, and kinematic open-source data for comprehensive locomotion
  activities,'' \emph{Sci. Data}, vol.~10, p. 789, 2023.
\bibitem{lazic2010pseudoreplication}
S.~E. Lazic, ``The problem of pseudoreplication in neuroscientific studies: is
  it affecting your analysis?'' \emph{BMC Neurosci.}, vol.~11, p.~5, 2010.
\bibitem{holm1979simple}
S.~Holm, ``A simple sequentially rejective multiple test procedure,''
  \emph{Scand. J. Stat.}, vol.~6, no.~2, pp. 65--70, 1979.
\bibitem{young2014training}
A.~J. Young, A.~M. Simon, and L.~J. Hargrove, ``A training method for
  locomotion mode prediction using powered lower limb prostheses,'' \emph{IEEE
  Trans. Neural Syst. Rehabil. Eng.}, vol.~22, no.~3, pp. 671--677, 2014.
\bibitem{muller2017twolearners}
J.~S. M{\"u}ller, C.~Vidaurre, M.~Schreuder, F.~C. Meinecke, P.~von B{\"u}nau,
  and K.-R. M{\"u}ller, ``A mathematical model for the two-learners problem,''
  \emph{J. Neural Eng.}, vol.~14, no.~3, p. 036005, 2017.
\end{thebibliography}

\end{document}